\documentclass[10pt,twocolumn,letterpaper]{article}

\usepackage{cvpr}                % To produce the CAMERA-READY version
\usepackage{graphicx}
\usepackage{amsmath}
\usepackage{amssymb}
\usepackage{booktabs}
\usepackage{multirow}
\usepackage{epstopdf}
\usepackage{bm}
\usepackage{booktabs}
\usepackage{url}
\usepackage{makecell}
\usepackage{cite}
\usepackage{color}
\usepackage{bbm}
\usepackage[pagebackref,breaklinks,colorlinks]{hyperref}

\usepackage[capitalize]{cleveref}
\crefname{section}{Sec.}{Secs.}
\Crefname{section}{Section}{Sections}
\Crefname{table}{Table}{Tables}
\crefname{table}{Tab.}{Tabs.}

\def\cvprPaperID{2729} % *** Enter the CVPR Paper ID here
\def\confName{CVPR}
\def\confYear{2022}

\begin{document}

%%%%%%%%% TITLE - PLEASE UPDATE
\title{FineDiving: A Fine-grained Dataset for Procedure-aware \\ Action Quality Assessment}

\author{
	Jinglin Xu\thanks{Equal contribution. ~\textsuperscript{$\dag$}Corresponding author.},
	Yongming Rao\footnotemark[1],
	Xumin Yu,
	Guangyi Chen,
	Jie Zhou,
	Jiwen Lu$^\dag$ \\
    {Department of Automation, Tsinghua University, China}\\
    {Beijing National Research Center for Information Science and Technology, China}\\
    {\tt\small \{xujinglinlove, raoyongming95\}@gmail.com; yuxm20@mails.tsinghua.edu.cn;} \\
    \vspace{-5pt}
    {\tt\small guangyichen1994@gmail.com; \{jzhou, lujiwen\}@tsinghua.edu.cn}
	}

% For a paper whose authors are all at the same institution,
% omit the following lines up until the closing ``}''.
% Additional authors and addresses can be added with ``\and'',
% just like the second author.
% To save space, use either the email address or home page, not both

\makeatletter
\let\@oldmaketitle\@maketitle
\renewcommand{\@maketitle}{\@oldmaketitle
\begin{minipage}{\textwidth}
\vspace{-5mm}
\centering
\includegraphics[width=0.96\linewidth]{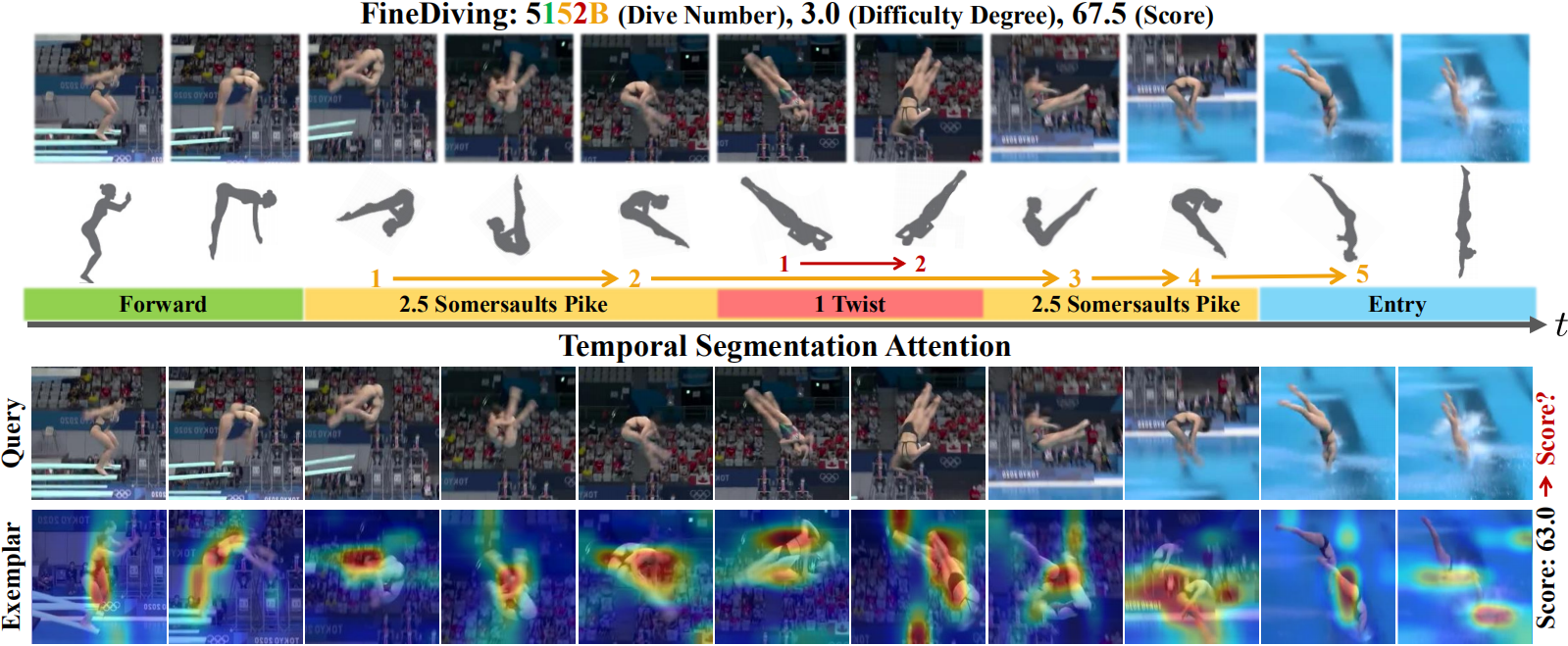}
\vspace{-3mm}
\captionof{figure}{An overview of the \textbf{FineDiving} dataset and procedure-aware action quality assessment approach. FineDiving is a fine-grained sports video dataset with detailed annotations on action procedures. It provides a potential for proposing an action quality assessment approach with better interpretability via constructing a new Temporal Segmentation Attention module between query and exemplar instances.
\label{top}}
\end{minipage}}
\makeatother

\maketitle

%%%%%%%%% ABSTRACT
\begin{abstract}
\vspace{-6pt}
  Most existing action quality assessment methods rely on the deep features of an entire video to predict the score, which is less reliable due to the non-transparent inference process and poor interpretability. We argue that understanding both high-level semantics and internal temporal structures of actions in competitive sports videos is the key to making predictions accurate and interpretable. Towards this goal, we construct a new fine-grained dataset, called FineDiving, developed on diverse diving events with detailed annotations on action procedures. We also propose a procedure-aware approach for action quality assessment, learned by a new Temporal Segmentation Attention module. Specifically, we propose to parse pairwise query and exemplar action instances into consecutive steps with diverse semantic and temporal correspondences. The procedure-aware cross-attention is proposed to learn embeddings between query and exemplar steps to discover their semantic, spatial, and temporal correspondences, and further serve for fine-grained contrastive regression to derive a reliable scoring mechanism. Extensive experiments demonstrate that our approach achieves substantial improvements over the state-of-the-art methods with better interpretability. The dataset and code are available at \url{https://github.com/xujinglin/FineDiving}.
\end{abstract}

%%%%%%%%% BODY TEXT
\vspace{-16pt}
\section{Introduction}
\label{sec:intro}

Competitive sports video understanding has become a hot research topic in the computer vision community. As one of the key techniques of understanding sports action, Action Quality Assessment (AQA) has attracted growing attention in recent years. In the 2020 Tokyo Olympic Games, the AI scoring system in gymnastics acted as a judge for assessing the athlete's score performance and providing feedback for improving the athlete's competitive skill, which reduces the controversies in many subjective scoring events, e.g., diving and gymnastics.

AQA is a task to assess how well an action is performed by estimating a score after analyzing the performance.
Unlike conventional action recognition \cite{schuldt2004recognizing,wang2013action,oneata2013action,ji20123d,simonyan2014two,tran2015learning,feichtenhofer2016convolutional,carreira2017quo,varol2017long,wang2018non,li2019action,yang2020temporal,shao2020finegym} and detection \cite{yeung2016end,montes2016temporal,zhao2017temporal,lin2019bmn}, AQA is more challenging since an action can be recognized from just one or a few images while the judges need to go through the entire action sequence to assess the action performance. Most existing AQA methods \cite{zhang2014relative,pirsiavash2014assessing,parmar2017learning,li2018end,doughty2018s,parmar2019action,mtlaqa,xu2019learning,pan2019action,tang2020uncertainty,bertasius2017baller,doughty2019pros,gattupalli2017cognilearn} regress on the deep features of videos to learn the diverse scores, which is difficult for actions with a small discrepancy happening in similar backgrounds. Since the diving events are usually filmed in a similar environment (i.e., aquatics centers) and all the videos contain the same action routine, that is ``take-off'', ``flight'', and ``entry'', while the subtle differences mainly appear in the numbers of both somersault and twist, flight positions as well as their executed qualities.
Capturing these subtle differences requires the AQA method not only to parse the steps of diving action but also to explicitly quantify the executed qualities of these steps. If we judge the action quality only via regressing a score on the deep features of the whole video, it would be a confusing and non-transparent assessment of the action quality, since we cannot explain the final score via analyzing the performances of action steps.

Cognitive science \cite{schmidt1976understanding,meyer2011assessing} shows that humans learn to assess the action quality by introducing fine-grained annotations and reliable comparisons. Inspired by this, we introduce these two concepts into AQA, which is challenging since existing AQA datasets lack fine-grained annotations of action procedures and cannot make reliable comparisons. If we judge the action quality using coarse-grained labels, we cannot date back to a convincing reason from the final action quality score. It is urgent to construct a fine-grained sports video dataset for encouraging a more transparent and reliable scoring approach for AQA.

To address these challenges, we construct a new competitive sports video dataset, ``FineDiving'' (short for Fine-grained Diving), focusing on various diving events, which is the first fine-grained sports video dataset for assessing action quality. FineDiving has several characteristics (Figure \ref{top}, the top half): (1) Two-level semantic structure. All videos are annotated with semantic labels at two levels, namely, action type and sub-action type, where a combination of the presented sub-action types produces an action type. (2) Two-level temporal structure. The temporal boundaries of actions in each video are annotated, where each action is manually decomposed into consecutive steps according to a well-defined lexicon. (3) Official action scores, judges' scores, and difficulty degrees are collected from FINA.

We further propose a procedure-aware approach for assessing action quality on FineDiving (Figure \ref{top}, the bottom half), inspired by the recently proposed CoRe \cite{yu2021group}. The proposed framework learns procedure-aware embeddings with a new Temporal Segmentation Attention module (referred to as TSA) to predict accurate scores with better interpretability. Specifically, TSA first parses action into consecutive steps with semantic and temporal correspondences, serving for procedure-aware cross-attention learning. The consecutive steps of query action are served as queries and the steps of exemplar action are served as keys and values. Then TSA inputs pairwise query and exemplar steps into the transformer and obtains procedure-aware embeddings via cross-attention learning. Finally, TSA performs fine-grained contrastive regression on the procedure-aware embeddings to quantify step-wise quality differences between query and exemplar, and predict the action score.

The contributions of this work are summarized as:
(1) We construct the first fine-grained sports video dataset for action quality assessment, which contains rich semantics and diverse temporal structures. (2) We propose a procedure-aware approach for action quality assessment, which is learned by a new temporal segmentation attention module and quantifies quality differences between query and exemplar in a fine-grained way. (3) Extensive experiments illustrate that our procedure-aware approach obtains substantial improvements and achieves the state-of-the-art.

%%%%%%%%%%%%%%%%%%%%%%%%%%%%%%%%%%%%%%%
\section{Related Work}
\noindent\textbf{Sports Video Datasets.}
Action understanding in sports videos is a hot research topic in the computer vision community, which is more challenging than understanding actions in the general video datasets, e.g., HMDB \cite{kuehne2011hmdb}, UCF-101 \cite{soomro2012ucf101}, Kinetics \cite{carreira2017quo}, AVA \cite{gu2018ava}, ActivityNet \cite{caba2015activitynet}, THUMOS \cite{THUMOS15}, Moments in Time \cite{monfortmoments} or HACS \cite{zhao2019hacs}, due to the low inter-class variance in motions and environments.
Competitive sports video action understanding relies heavily on available sports datasets.
Early, Niebles \textit{et al.} \cite{niebles2010modeling} introduced the Olympic sports dataset into modeling the motions. Karpathy \textit{et al.} \cite{karpathy2014large} provided a large-scale dataset Sports1M and gained significant performance over strong feature-based baselines. Pirsiavash \textit{et al.} \cite{pirsiavash2014assessing} released the first Olympic judging dataset comprising of Diving and Figure Skating. Parmar \textit{et al.} \cite{parmar2017learning} released a new dataset including Diving, Gymnastic Vault, Figure Skating for better working on AQA. Bertasius \textit{et al.} \cite{bertasius2017baller} proposed a first-person basketball dataset for estimating performance assessment of basketball players. Li \textit{et al.} \cite{li2018resound} built the Diving48 dataset annotated by the combinations of 4 attributes (i.e., back, somersault, twist, and free). Xu \textit{et al.} \cite{xu2019learning} extended the existing MIT-Figure Skating dataset to 500 samples. Parmar \textit{et al.} \cite{mtlaqa,parmar2019action} presented the MTL-AQA dataset that exploits multi-task networks to assess the motion. Shao \textit{et al.} \cite{shao2020finegym} proposed the FineGym dataset that provides coarse-to-fine annotations both temporally and semantically for facilitating action recognition.
Recently, Li \textit{et al.} \cite{li2021multisports} developed a large-scale dataset MultiSports with fine-grained action categories with dense annotations in both spatial and temporal domains for spatio-temporal action detection. Hong \textit{et al.} \cite{hong2021video} provided a figure skating dataset VPD for facilitating fine-grained sports action understanding. Chen \textit{et al.} \cite{chen2021sportscap} proposed the SMART dataset with fine-grained semantic labels, 2D and 3D annotated poses, and assessment information.
Unlike the above datasets, FineDiving is the first fine-grained sports video dataset for AQA, which guides the model to understand action procedures via detailed annotations towards more reliable action quality assessment.

\begin{figure}[t]
  \centering
  \includegraphics[width=\linewidth]{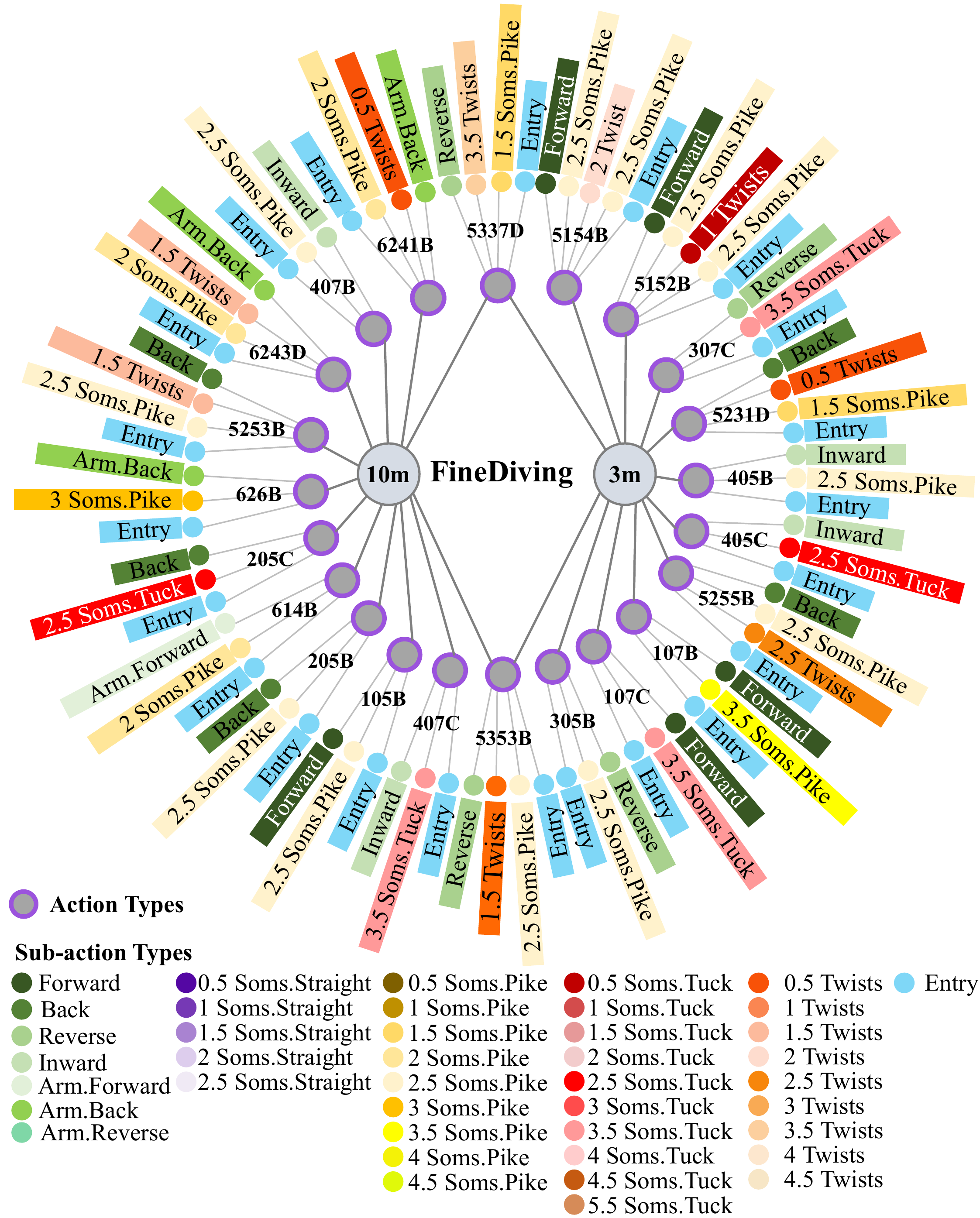}
  \caption{Two-level semantic structure. Action type indicates an action routine described by a dive number. Sub-action type is a component of action type, where each combination of the sub-action types can produce an action type and different action types can share the same sub-action type. The green branch denotes different kinds of take-offs. The purple, yellow, and red branches respectively represent the somersaults with three positions (i.e., straight, pike, and tuck) in the flight, where each branch contains different somersault turns. The orange branch indicates different twist turns interspersed in the process of somersaults. The light blue denotes entering the water. (Best viewed in color.)}
  \label{semantic}
  \vspace{-10pt}
\end{figure}

\begin{figure}[t]
  \centering
  \includegraphics[width=\linewidth]{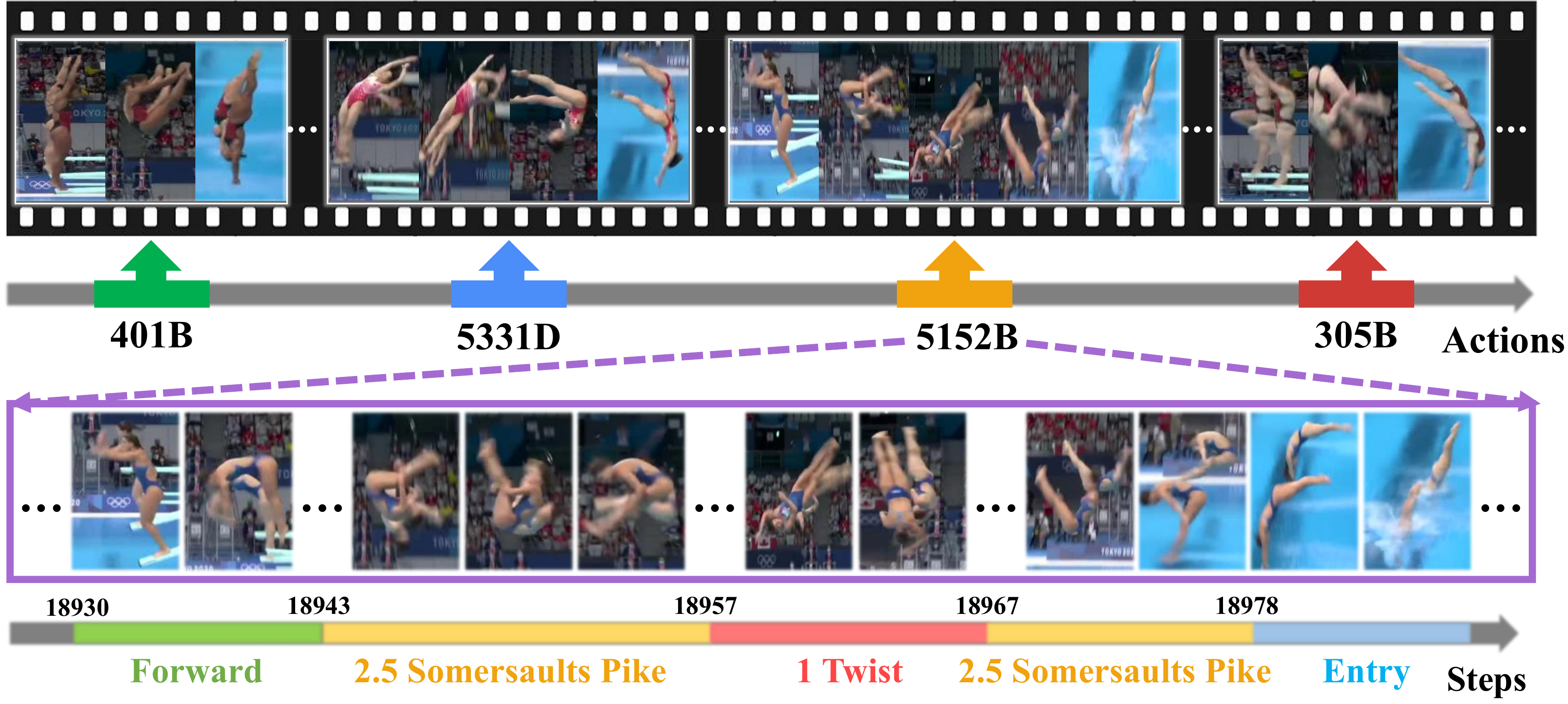}
  \caption{Two-level temporal structure. The action-level labels describe temporal boundaries of valid action routines, while the step-level labels provide the starting frames of consecutive steps in the procedure. (Best viewed in color.)}
  \label{temporal}
  \vspace{-10pt}
\end{figure}

\noindent\textbf{Action Quality Assessment.}
Most existing methods formulate AQA as a regression on various video representations supervised by the scores. In early pioneering work, Pirsiavash \textit{et al.} \cite{pirsiavash2014assessing} first formulated AQA and proposed to map the pose-based features, spatio-temporal interest points, and hierarchical convolutional features to the scores by using SVR. Parmar \textit{et al.} \cite{parmar2017learning} utilized the spatio-temporal features to estimate scores and demonstrated its effectiveness on actions like Diving, Gymnastic Vault, and Figure Skating. Bertasius \textit{et al.} \cite{bertasius2017baller} proposed a learning-based approach to estimate motion, behaviors, and performance assessment of basketball players. Li \textit{et al.} \cite{li2018end} proposed to combine some network modification with ranking loss to improve the AQA performance. Doughty \textit{et al.} \cite{doughty2019pros} assessed the relative overall level of skill in a long video based on the video-level pairwise annotation via the high-skill and low-skill attention modules. Parmar \textit{et al.} \cite{parmar2019action} introduced the shared concepts of action quality among actions. Parmar \textit{et al.} \cite{mtlaqa} further reformulated the definition of AQA as a multitask learning in an end-to-end fashion. Besides, Xu \textit{et al.} \cite{xu2019learning} proposed to use self-attentive and multiscale skip convolutional LSTM to aggregate information from individual clips, which achieved the best performance on the assessment of Figure Skating samples. Pan \textit{et al.} \cite{pan2019action} assessed the performance of actions visually from videos by graph-based joint relation modeling.
Recently, Tang \textit{et al.} \cite{tang2020uncertainty} proposed to reduce the underlying ambiguity of the action score labels from human judges via an uncertainty-aware score distribution learning (USDL). Yu \textit{et al.} \cite{yu2021group} constructed a contrastive regression framework (CoRe) based on the video-level features to rank videos and predict accurate scores.
Different from previous methods, our approach understands action procedures and mines procedure-aware attention between query and exemplar to achieve a more transparent action assessment.

%%%%%%%%%%%%%%%%%%%%%%%%%%%%%%%%%%%%%%%
\section{The FineDiving Dataset}\label{findiving}
In this section, we propose a new fine-grained competitive sports video dataset FineDiving. We will introduce FineDiving from dataset construction and statistics.

\subsection{Dataset Construction}
\noindent\textbf{Collection.}
We search for diving events in Olympics, World Cup, World Championships, and European Aquatics Championships on YouTube, and download competition videos with high-resolution. Each official video provides rich content, including diving records of all athletes and slow playbacks from different viewpoints.

\noindent\textbf{Lexicon.}
We construct a fine-grained video dataset organized by both semantic and temporal structures, where each structure contains two-level annotations, shown in Figures \ref{semantic} and \ref{temporal}. Herein, we employ three professional athletes of the diving association, who have prior knowledge in diving and help to construct a lexicon for subsequent annotation.

For semantic structure in Figure \ref{semantic}, the action-level labels describe the action types of athletes and the step-level labels depict the sub-action types of consecutive steps in the procedure, where adjacent steps in each action procedure belong to different sub-action types. A combination of sub-action types produces an action type.  For instance, for an action type ``5255B'', the steps belonging to the sub-action types ``Back'', ``2.5 Somersaults Pike'', and ``2.5 Twists'' are executed sequentially.

In temporal structure, the action-level labels locate the temporal boundary of a complete action instance performed by an athlete. During this annotation process, we discard all the incomplete action instances and filter out the slow playbacks. The step-level labels are the starting frames of consecutive steps in the action procedure. For example, for an action belonging to the type ``5152B'', the starting frames of consecutive steps are 18930, 18943, 18957, 18967, and 18978, respectively, shown in Figure \ref{temporal}.

\begin{figure}[t]
  \centering
  \includegraphics[width=\linewidth]{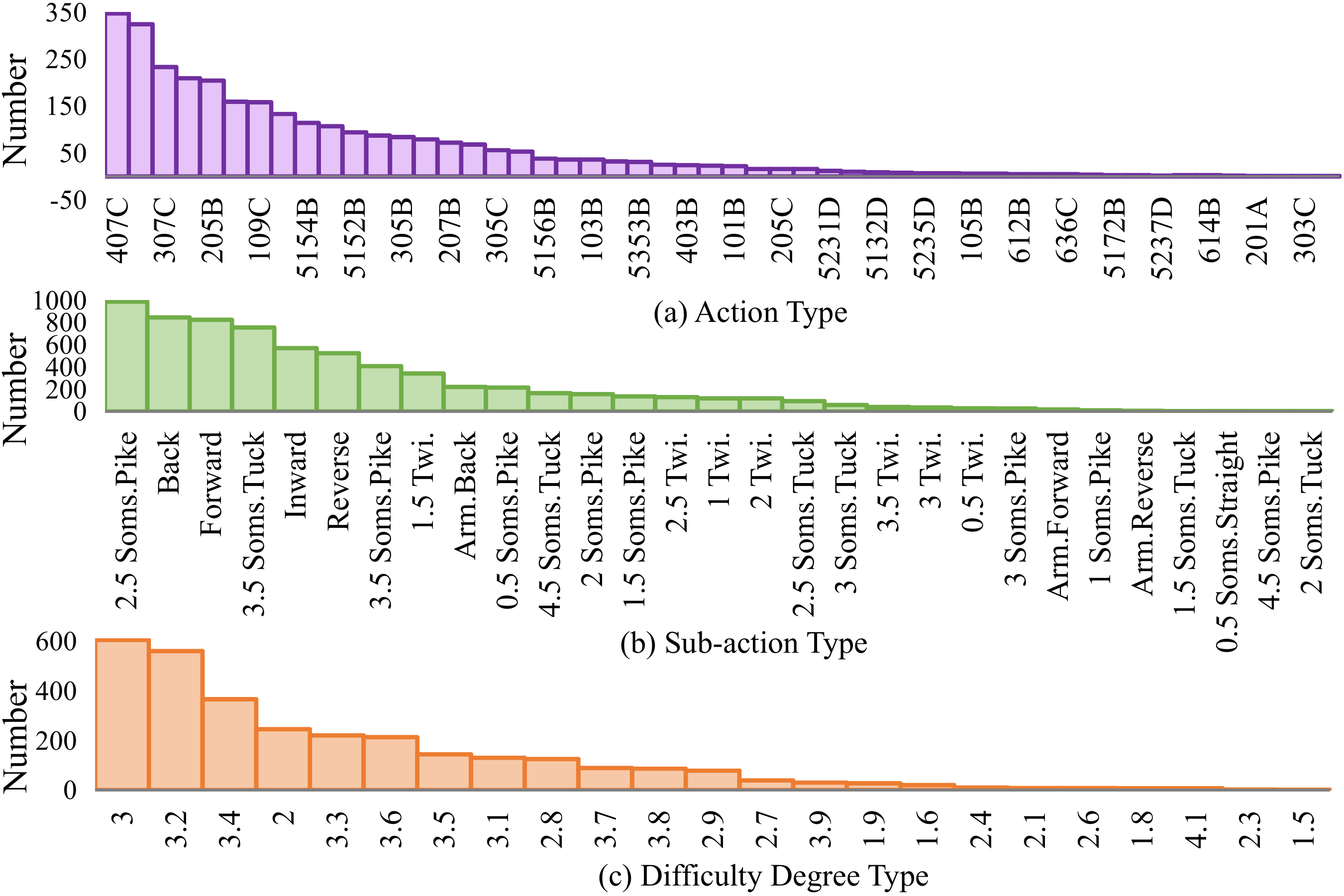}
  \caption{Statistics of FineDiving. (a) The action-type distribution of action instances. (b) The sub-action type distribution of action instances. (c) The difficulty degree distribution of action instances.}
  \label{statistic}
  \vspace{-10pt}
\end{figure}

\noindent\textbf{Annotation.}
Given a raw diving video, the annotator utilizes our defined lexicon to label each action and its procedure. We need to accomplish two annotation stages from coarse- to fine-grained. The coarse-grained stage is to label the action type for each action instance and its temporal boundary accompanied with the official score. The fine-grained stage is to label the sub-action type for each step in the action procedure and record the starting frame of each step. Both coarse- and fine-grained annotation stages adopt a cross-validating labeling method. Specifically, we employ six workers who have prior knowledge in the diving domain and divide data into six parts without overlap. The annotation results of one worker are checked and adjusted by another, which ensures annotation results are double-checked. To improve the annotation efficiency, we utilize an effective toolbox \cite{tang2019coin} in the fine-grained annotation stage. Under this pipeline, the total time of the whole annotation process is about 120 hours.

\subsection{Dataset Statistics}

The FineDiving dataset consists of 3000 video samples, covering 52 action types, 29 sub-action types, and 23 difficulty degree types, which are shown in Figure \ref{statistic}. These statistics will be helpful to design competition strategy and better bring athletes' superiority into full play. Table \ref{table:dataset} reports more detailed information on our dataset and compares it with existing AQA datasets as well as other fine-grained sports datasets. Our dataset is different from existing AQA datasets in the annotation type and dataset scale. For instance, MIT-Dive, UNLV, and AQA-7-Dive only provide action scores, while our dataset provides fine-grained annotations including action types, sub-action types, coarse- and fine-grained temporal boundaries as well as action scores. MTL-AQA provides coarse-grained annotations, i.e., action types and temporal boundaries. Other fine-grained sports datasets cannot be used for assessing action quality due to a lack of action scores. We see that FineDiving is the first fine-grained sports video dataset for the AQA task, filling the fine-grained annotations void in AQA.

\begin{table}[tb]
\renewcommand\tabcolsep{0.5pt}
\footnotesize
\centering
\caption{Comparison of existing sports video datasets and FineDiving. \textit{Score} indicates the score annotations; \textit{Step} is fine-grained class and temporal boundary; \textit{Action} is coarse-grained class and temporal boundary; \textit{Tube} contains fine-grained class, temporal boundary, and spatial localization.}
\vspace{-5pt}
\begin{tabular}{|l|c|c|c|c|c|}
\hline
Localization & \#Samples & \#Events & \#Act. Clas. & Avg.Dur. & Anno.Type \\
\hline
TAPOS\cite{shao2020intra} & 16294 & / & 21 & 9.4s & Step \\
FineGym\cite{shao2020finegym} & 32697 & 10 & 530 & 1.7s& Step \\
MultiSports\cite{li2021multisports} & 37701 & 247 & 66 & 1.0s & Tube \\
\hline
Assessment & \#Samples & \#Events & \#Sub-act.Typ. & Avg.Dur. & Anno.Type \\
\hline
MIT Dive\cite{pirsiavash2014assessing} & 159  &  1  & / & 6.0s & Score \\
UNLV Dive\cite{parmar2017learning} & 370  &  1  & /  & 3.8s & Score \\
AQA-7-Dive\cite{parmar2019action} & 549 &  6 & / & 4.1s  & Score \\
MTL-AQA\cite{mtlaqa} & 1412 & 16   &  / & 4.1s & Action,Score\\
\hline
\textbf{FineDiving} & 3000  & 30 & 29 & 4.2s & Step,Score\\
\hline
\end{tabular}
\label{table:dataset}
\vspace{-10pt}
\end{table}

\begin{figure*}[t]
  \centering
  \includegraphics[width=0.96\linewidth]{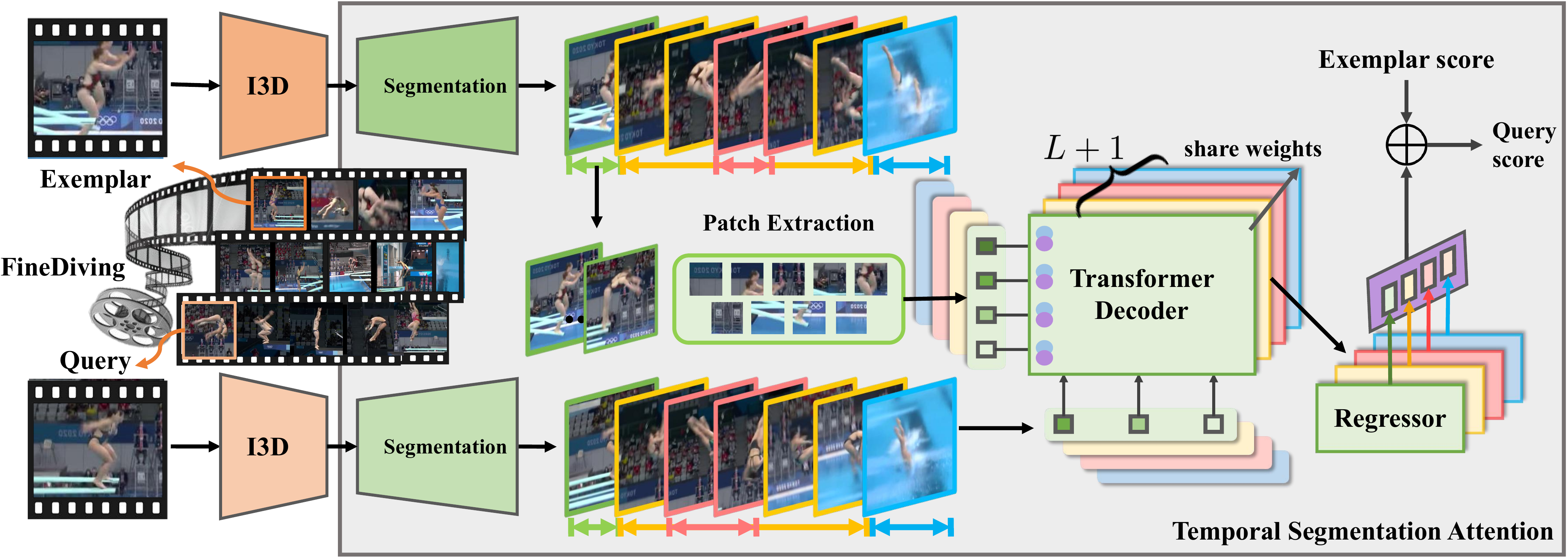}
  \caption{The architecture of the proposed procedure-aware action quality assessment. Given a pairwise query and exemplar instances, we extract spatial-temporal visual features with I3D and propose a Temporal Segmentation Attention module to assess action quality via successively accomplishing procedure segmentation, procedure-aware cross-attention learning, and fine-grained contrastive regression. The temporal segmentation attention module is supervised by step transition labels and action score labels, which guides the model to focus on exemplar regions that are consistent with the query step and quantify their differences to predict reliable action scores.}
  \label{framework}
  \vspace{-10pt}
\end{figure*}

\section{Approach}\label{our_approach}
In this section, we will systematically introduce our approach, whose main idea is to construct a new temporal segmentation attention module to propose a reliable and transparent action quality assessment approach. The overall architecture of our approach is illustrated in Figure \ref{framework}.

\subsection{Problem Formulation}
Given pairwise query $X$ and exemplar $Z$ instances, our procedure-aware approach is formulated as a regression problem that predicts the action quality score of the query video via learning a new Temporal Segmentation Attention module (abbreviated as TSA). It can be represented as:
\begin{equation}
    \hat{y}_X=\mathcal{P}(X,Z|\Theta)+y_Z
\end{equation}
where $\mathcal{P}\!=\!\{\mathcal{F},\mathcal{T}\}$ denotes the overall framework containing I3D \cite{carreira2017quo} backbone $\mathcal{F}$ and TSA module $\mathcal{T}$; $\Theta$ indicates the learnable parameters of $\mathcal{P}$; $\hat{y}_X$ is the predicted score of $X$ and $y_Z$ is the ground-truth score of $Z$.

\subsection{Temporal Segmentation Attention}
There are three components in TSA, that is, procedure segmentation, procedure-aware cross-attention learning, and fine-grained contrastive regression.

\noindent\textbf{Procedure Segmentation.} To parse pairwise query and exemplar actions into consecutive steps with semantic and temporal correspondences, we first propose to segment the action procedure by identifying the transition in time that the step switches from one sub-action type to another.

Suppose that $L$ step transitions are needed to be identified, the procedure segmentation component $\mathcal{S}$ predicts the probability of the step transition occurring at the $t$-th frame by computing:
\begin{align}
&[\hat{p}_1,\cdots,\hat{p}_L]=\mathcal{S}(\mathcal{F}(X)), \label{PS}\\
&\hat{t}_k=\underset{\frac{T}{L}(k-1)< t\leq\frac{T}{L}k}{\arg\max}\ \hat{p}_k(t) \label{pred_t}
\end{align}
where $\hat{p}_k\in\mathbb{R}^{T}$ is the predicted probability distribution of the $k$-th step transition; $\hat{p}_k(t)$ denotes the predicted probability of the $k$-th step transiting at the $t$-th frame; $\hat{t}_k$ is the prediction of the $k$-th step transition.

In Equation (\ref{PS}), the component $\mathcal{S}$ is composed of two blocks, namely ``down-up'' ($b_1$) and ``linear'' ($b_2$).
Specifically, the $b_1$ block consists of four ``down-$m$-up-$n$'' sub-blocks, where $m$ and $n$ denote specified dimensions of output along the spatial and temporal axes, respectively. Each sub-block contains two consecutive convolution layers and one max pooling layer. The $b_1$ block increases the length of I3D features $\mathcal{F}(X)$ using convolution layers along the temporal axis and reduces the dimension of $\mathcal{F}(X)$ using max-pooling layers along the spatial axis. The ``down-up'' block advances the visual features $\mathcal{F}(X)$ in deeper layers to contain both deeper spatial and longer temporal views for procedure segmentation. The ``linear'' block further encodes the output of the $b_1$ block to generate $L$ probability distributions $\{\hat{p}_k\}_{k=1}^L$ of $L$ step transitions in an action procedure. Besides, the constrain in Equation (\ref{pred_t}) ensures the predicted transitions being ordered, i.e., $\hat{t}_1\leq\cdots\leq\hat{t}_L$.

Given the ground-truth of the $k$-th step transition, i.e., $t_k$, it can be encoded as a binary distribution $p_k$, where $p_k(t_k)\!=\!1$ and $p_k(t_s)|_{s\neq k}\!=\!0$. With the prediction $\hat{p}_k$ and ground-truth $p_k$, the procedure segmentation problem can be converted to a dense classification problem, which predicts the probability of whether each frame is the $k$-th step transition.
We calculate the binary cross-entropy loss between $\hat{p}_k$ and $p_k$ to optimize $\mathcal{S}$ and find the frame with the greatest probability of being the $k$-th step transition. The objective function can be written as:
\begin{equation}
\begin{split}
\mathcal{L}_{\text{BCE}}(\hat{p}_k, p_k)&=-\textstyle\sum_t(p_k(t)\log \hat{p}_k(t) \\
                                       &+(1-p_k(t))\log (1-\hat{p}_k(t))).
\end{split}
\end{equation}
Minimizing $\mathcal{L}_{\text{BCE}}$ makes distributions $\hat{p}_k$ and $p_k$ closer.
\vspace{1pt}

\noindent\textbf{Procedure-aware Cross-Attention.}
Through procedure segmentation, we obtain $L\!+\!1$ consecutive steps with semantic and temporal correspondences in each action procedure based on $L$ step transition predictions. We leverage the sequence-to-sequence representation ability of the transformer for learning procedure-aware embedding of pairwise query and exemplar steps via cross-attention.

Based on $\mathcal{S}$, the query and exemplar action instances are divided into $L\!+\!1$ consecutive steps, denoted as $\{S^X_l,S^Z_l\}_{l=1}^{L+1}$. Considering that the lengths of $S^X_l$ and $S^Z_l$ may be different, we fix them into the given size via downsampling or upsampling to meet the requirement that the dimensions of ``query'' and ``key'' are the same in the attention model. Then we propose procedure-aware cross-attention learning to discover the spatial and temporal correspondences between pairwise steps $S^X_l$ and $S^Z_l$, and generate new features in both of them. The pairwise steps complement each other and guide the model to focus on the consistent region in $S^Z_l$ with $S^X_l$. Here, $S^Z_l$ preserves some spatial information from the feature map (intuitively represented by patch extraction in Figure \ref{framework}). The above procedure-aware cross-attention learning can be represented as:
\begin{align}
S_l^{r'}&\!=\!\text{MCA}(\text{LN}(S_l^{r-1},S_l^Z))\!+\!S_l^{r-1},&r\!=\!1\!\cdots\!R\\
S_l^r&\!=\!\text{MLP}(\text{LN}(S_l^{r'}))\!+\!S_l^{r'},&r\!=\!1\!\cdots\!R
\end{align}
where $S_l^0\!=\!S_l^X$, $S_l\!=\!S_l^R$. The transformer decoder \cite{dosovitskiy2020image} consisted of alternating layers of Multi-head Cross-Attention (MCA) and MLP blocks, where the LayerNorm (LN) and residual connections are applied before and after every block, respectively, and the MLP block contains two layers with a GELU non-linearity.

\noindent\textbf{Fine-grained Contrastive Regression.}
Based on the learned procedure-aware embedding $S_l$, we quantify the step deviations between query and exemplar by learning the relative scores of pairwise steps, which guides the TSA module to assess action quality via learning fine-grained contrastive regression component $\mathcal{R}$.
It is formulated as:
\begin{equation}
    \hat{y}_X = \frac{1}{L+1}\sum_{l=1}^{L+1}\mathcal{R}(S_l) + y_Z
\end{equation}
where $y_Z$ is the exemplar score label from the training set.
We optimize $\mathcal{R}$ by computing the mean squared error between the ground truth $y_X$ and prediction $\hat{y}_X$, that is:
\begin{equation}
    \mathcal{L}_{\text{MSE}}=\|\hat{y}_X-y_X\|^2.
\end{equation}

\subsection{Optimization and Inference}
During training, for pairwise query and exemplar $(X,Z)$ in the training set with step transition labels and action score labels,
the final objective function for the video pair is:
\begin{equation}
    J=\sum_{k=1}^L\mathcal{L}_{\text{BCE}}(\hat{p}_k, p_k)+\mathcal{L}_{\text{MSE}}.
\end{equation}

During testing, for a test video $X_{\text{test}}$, we adopt a multi-exemplar voting strategy \cite{yu2021group} to select $M$ exemplars from the training set and then construct $M$ video pairs $\{(X_{\text{test}},Z_j)\}_{j=1}^M$ with exemplar score labels $\{y_{Z_j}\}_{j=1}^M$.
The process of multi-exemplar voting can be written as:
\begin{equation}
    \hat{y}_{X_{\text{test}}}=\frac{1}{M}\textstyle\sum_{j=1}^M(\mathcal{P}(X_{\text{test}},Z_j|\Theta)+y_{Z_j}).
\end{equation}

\section{Experiments}
\subsection{Evaluation Metrics}
We comprehensively evaluate our approach on two aspects, namely procedure segmentation and action quality assessment, and compute the following three metrics.

\noindent\textbf{Average Intersection over Union.}
When the procedure segmentation is finished, a set of predicted step transitions are obtained for each video sample.
We rewrite these step transition predictions as a set of 1D bounding boxes, denoted as $\mathcal{B}_p\!=\!\{\hat{t}_{k+1}\!-\!\hat{t}_k\}_{k=1}^{L-1}$. Supposed that the ground-truth bounding boxes can be written as $\mathcal{B}_g\!=\!\{t_{k+1}\!-\!t_k\}_{k=1}^{L-1}$, we calculate the average Intersection over Union (AIoU) between two bounding boxes (i.e., $\hat{t}_{k+1}\!-\!\hat{t}_k$ and $t_{k+1}\!-\!t_k$) and determine the correctness of each prediction if IoU$_i$ is larger than a certain threshold $d$.
We describe the above operation as a metric AIoU$@d$ for evaluating our approach:
\begin{align}
    \text{AIoU}@d&=\frac{1}{N}\textstyle\sum_{i=1}^N\mathcal{I}(\text{IoU}_i\geq d)\\
    \text{IoU}_i&=|\mathcal{B}_p\cap\mathcal{B}_g|/|\mathcal{B}_p\cup\mathcal{B}_g|
\end{align}
where $\text{IoU}_i$ indicates the Intersection over Union for the $i$-th sample and $\mathcal{I}(\cdot)$ is an indicator that outputs $1$ if $\text{IoU}_i\!\geq\!d$, whereas outputs $0$. The higher of AIoU$@d$, the better of procedure segmentation.

\noindent\textbf{Spearman's rank correlation.} Following the previous work \cite{parmar2019action,pan2019action,mtlaqa,tang2020uncertainty,yu2021group}, we adopt Spearman's rank correlation ($\rho$) to measure the AQA performance of our approach. $\rho$ is defined as:
\begin{equation}
    \rho=\frac{\textstyle\sum_{i=1}^N(y_i-\bar{y})(\hat{y}_i-\bar{\hat{y}})}{\sqrt{\textstyle\sum_{i=1}^N(y_i-\bar{y})^2\textstyle\sum_{i=1}^N(\hat{y}_i-\bar{\hat{y}})^2}}
\end{equation}
where $\boldsymbol{y}$ and $\boldsymbol{\hat{y}}$ denote the ranking of two series, respectively. The higher $\rho$ the better performance.

\noindent\textbf{Relative $\ell_2$-distance.} Following \cite{yu2021group}, we also utilize relative $\ell_2$-distance (R-$\ell_2$) to measure the AQA performance of our approach. Given the highest and lowest scores of an action, namely $y_{\max}$ and $y_{\min}$, R-$\ell_2$ can be defined as:
\begin{equation}
    \text{R-}\ell_2=\frac{1}{N}\sum_{i=1}^N
    \left(\frac{|y_i-\hat{y}_i|}{y_{\max}-y_{\min}}\right)
\end{equation}
where $y_i$ and $\hat{y}_i$ indicate the ground-truth and predicted scores for the $i$-th sample, respectively. The lower of $R_{\ell_2}$, the better performance.

\subsection{Implementation Details}
\noindent\textbf{Experiment Settings.}
We adopted the I3D model pre-trained on the Kinetics \cite{carreira2017quo} dataset as $\mathcal{F}$ with the initial learning rate 10$^{-4}$. We set the initial learning rates of $\mathcal{T}$ as 10$^{-3}$. We utilized Adam \cite{kingma2014adam} optimizer and set weight decay as 0. Similar to \cite{tang2020uncertainty,yu2021group}, we extracted 96 frames for each video, split them into 9 snippets, and then fed them into I3D, where each snippet contains 16 continuous frames with stride 10 frames. Following the experiment settings in \cite{mtlaqa,tang2020uncertainty,yu2021group}, we selected 75 percent of samples are for training and 25 percent are for testing in all the experiments. We also specified network parameters for two blocks $b_1$ and $b_2$ in $\mathcal{S}$. In the block $b_1$, ($m,n$) in the sub-blocks equal to (1024, 12), (512, 24), (256, 48), and (128, 96), respectively. The block $b_2$ is a three-layer MLP.
Furthermore, we set $M$ as 10 in the multi-exemplar voting strategy and set the number of step transitions $L$ as 2. More details about the criterion of selecting exemplars and the number of step transitions can be found in the supplementary materials.

\begin{table}[t]
\caption{Comparisons of performance with existing AQA methods on FineDiving. (w/o DN) indicates selecting exemplars randomly; (w/ DN) indicates using dive numbers to select exemplars; / indicates without procedure segmentation.}
\vspace{-5pt}
\footnotesize
\renewcommand\tabcolsep{6pt}
\label{table:results}
\centering
\begin{tabular}{l|cc|c|c}
\toprule
\multirow{2}{*}{Method (w/o DN)} &\multicolumn{2}{c|}{AIoU@} & \multirow{2}{*}{$\rho$} & \multirow{2}{*}{R-$\ell_2(\times100)$}  \\
\cline{2-3}
& 0.5 & 0.75 & & \\
\midrule
USDL \cite{tang2020uncertainty} & /  &  /  & 0.8302 & 0.5927\\
MUSDL \cite{tang2020uncertainty} & /  &  /  & 0.8427 & 0.5733\\
CoRe \cite{yu2021group} & /  &  / & 0.8631  & 0.5565\\
\midrule
\textbf{TSA} & \textbf{80.71} & \textbf{30.17} & \textbf{0.8925}  & \textbf{0.4782} \\
\bottomrule
\multirow{2}{*}{Method (w/ DN)} &\multicolumn{2}{c|}{AIoU@} & \multirow{2}{*}{$\rho$} & \multirow{2}{*}{R-$\ell_2(\times100)$}  \\
\cline{2-3}
& 0.5 & 0.75 & & \\
\midrule
USDL \cite{tang2020uncertainty} & /  &  /  & 0.8913  & 0.3822 \\
MUSDL \cite{tang2020uncertainty} & /  &  /  & 0.8978 & 0.3704\\
CoRe \cite{yu2021group} & /  &  / & 0.9061  & 0.3615 \\
\midrule
\textbf{TSA} & \textbf{82.51} & \textbf{34.31} &  \textbf{0.9203} &  \textbf{0.3420}\\
\bottomrule
\end{tabular}
\vspace{-10pt}
\end{table}

\noindent\textbf{Compared Methods.}
We reported the performance of the following methods including baseline and different versions of our approach:

$\bullet$ $\mathcal{F}$+$\mathcal{R}$ (Baseline), $\mathcal{F}$+$\mathcal{R}^\star$, and $\mathcal{F}$+$\mathcal{R}^\sharp$: The baseline uses I3D to extract visual features for each input video and predicts the score through a three-layer MLP with ReLU non-linearity, which is optimized by the MSE loss between the prediction and the ground truth.
$\star$ indicates the baseline adopting the asymmetric training strategy and $\sharp$ denotes the baseline concatenating dive numbers to features.

$\bullet$ $\mathcal{F}$+$\mathcal{S}$+$\mathcal{R}$: The procedure segmentation component is introduced into the baseline, which is optimized by the combination of MSE and BCE losses.

$\bullet$ TSA, TSA$^\dagger$: The approach was proposed in Section \ref{our_approach}. $\dagger$ indicates parsing action procedure using the ground-truth step transition labels instead of the prediction of procedure segmentation, which can be seen as an oracle for TSA.

\subsection{Results and Analysis}
\noindent\textbf{Comparisons with the State-of-the-art Methods.}
Table \ref{table:results} shows the experimental results of our approach and other AQA methods, trained and evaluated on the FineDiving dataset. We see that our approach achieves the state-of-the-art. Specifically, compared with the methods (USDL, MUSDL, and CoRe) without using dive numbers to select exemplars (i.e., w/o DN), our approach respectively obtained 6.23\%, 4.98\% and 2.94\% improvements on Spearman's rank correlation. Meanwhile, our approach also achieved 0.1145, 0.0951, and 0.0783 improvements on Relative $\ell_2$-distance compared to those methods, respectively.
Similarly, compared with USDL, MUSDL, and CoRe that use dive numbers to select exemplars (i.e., w/ DN), our approach also obtained different improvements on both Spearman's rank correlation and Relative $\ell_2$-distance.

\begin{table}[t]
\caption{Ablation studies on FineDiving. / indicates the methods without segmentation and $\checkmark$ denotes the method using the ground-truth step transition labels.}
\vspace{-5pt}
\footnotesize
\renewcommand\tabcolsep{6pt}
\label{table:ablation}
\centering
\begin{tabular}{l|cc|c|c}
\toprule
\multirow{2}{*}{Method (w/o DN)} & \multicolumn{2}{c|}{AIoU@} & \multirow{2}{*}{$\rho$} & \multirow{2}{*}{R-$\ell_2(\times100)$}\\
\cline{2-3}
& 0.5 & 0.75 & & \\
\midrule
$\mathcal{F}$+$\mathcal{R}$  & / & / & 0.8504 & 0.5837\\
$\mathcal{F}$+$\mathcal{R}^\star$ & / & / & 0.8452 & 0.6022\\
$\mathcal{F}$+$\mathcal{R}^\sharp$  & / & / & 0.8516 & 0.5736\\
$\mathcal{F}$+$\mathcal{S}$+$\mathcal{R}$ & 77.44 & 26.36 & 0.8602 & 0.5687\\
\textbf{TSA} & \textbf{80.71} & \textbf{30.17} & \textbf{0.8925}  & \textbf{0.4782} \\
TSA$^\dagger$ & $\checkmark$ & $\checkmark$ & 0.9029 & 0.4536 \\
\bottomrule
\multirow{2}{*}{Method (w/ DN)} & \multicolumn{2}{c|}{AIoU@} & \multirow{2}{*}{$\rho$} & \multirow{2}{*}{R-$\ell_2(\times100)$}\\
\cline{2-3}
& 0.5 & 0.75 & & \\
\midrule
$\mathcal{F}$+$\mathcal{R}$  & / & / & 0.8576 & 0.5695\\
$\mathcal{F}$+$\mathcal{R}^\star$ & / & / & 0.8563 & 0.5770\\
$\mathcal{F}$+$\mathcal{R}^\sharp$  & / & / & 0.8721 & 0.5435\\
$\mathcal{F}$+$\mathcal{S}$+$\mathcal{R}$ & 78.64 & 29.37 & 0.8793 & 0.5428 \\
\textbf{TSA} & \textbf{82.51} & \textbf{34.31} &  \textbf{0.9203} &  \textbf{0.3420}\\
TSA$^\dagger$ & $\checkmark$ & $\checkmark$ & 0.9310 & 0.3260\\
\bottomrule
\end{tabular}
\vspace{-10pt}
\end{table}

\begin{figure*}
  \centering
  \includegraphics[width=0.96\linewidth]{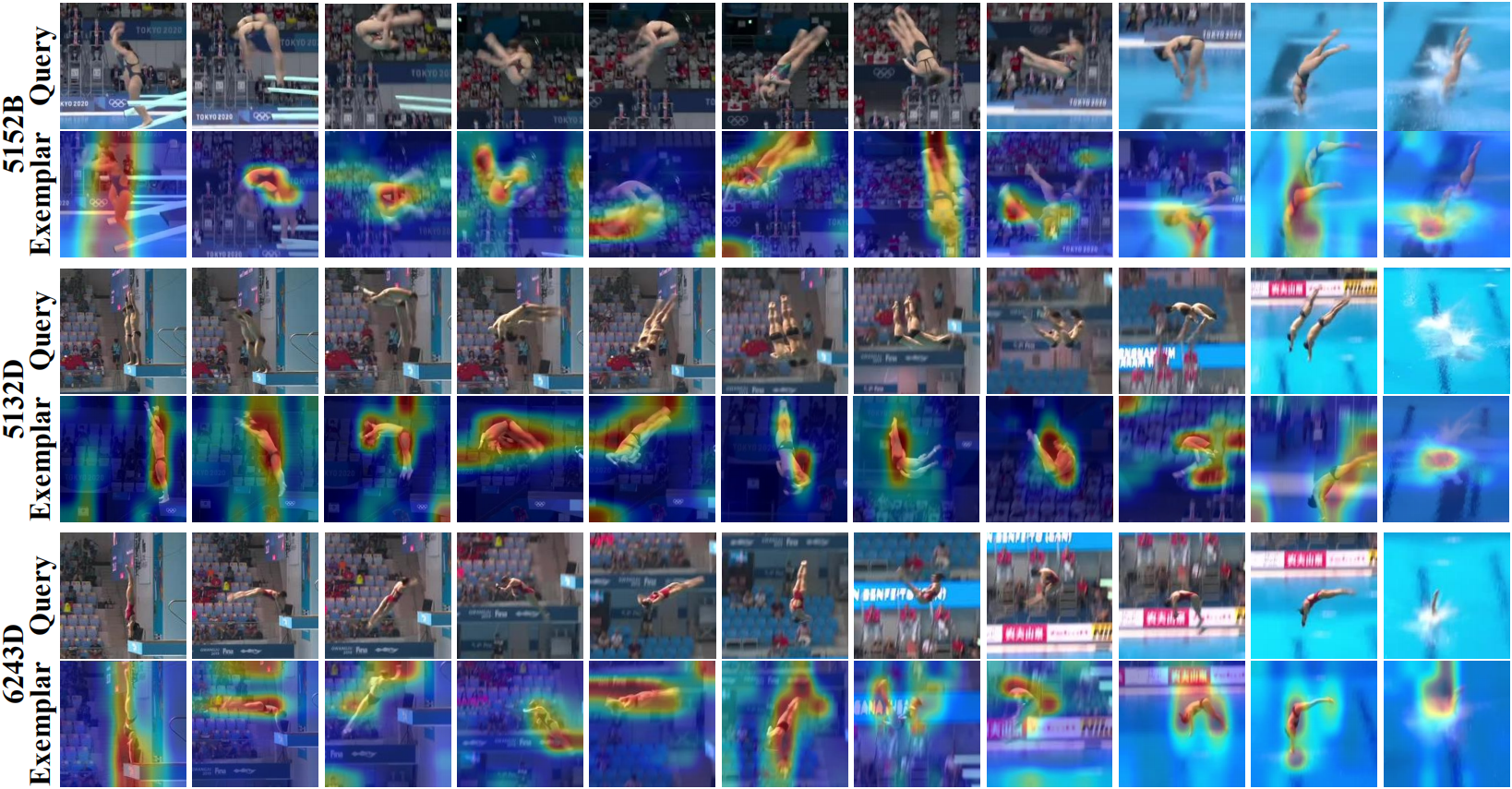}
  \vspace{-3pt}
  \caption{The visualization of procedure-aware cross attention between pairwise query and exemplar procedures. Our approach can focus on the exemplar regions that are consistent with the query step, which makes step-wise quality differences quantifying reliable. The presented pairwise query and exemplar contain the same action and sub-action types. (Best viewed in color.)}
  \label{visualization}
  \vspace{-12pt}
\end{figure*}

\noindent\textbf{Ablation studies.} We conducted some analysis experiments under two method settings for studying the effects of dive number, asymmetric training strategy, and procedure-aware cross-attention learning. As shown in Table \ref{table:ablation}, the performance of $\mathcal{F}$+$\mathcal{R}$ was slightly better than that of $\mathcal{F}$+$\mathcal{R}^\star$, which verified the effectiveness of symmetric training strategy. Compared with $\mathcal{F}$+$\mathcal{R}$, $\mathcal{F}$+$\mathcal{R}^\sharp$ obtained 0.12\% and 1.45\% improvements on Spearman's rank correlation. Meanwhile, our approach also achieved 0.0101 and 0.026 improvements on Relative $\ell_2$-distance. It demonstrated that concatenating dive numbers and I3D features can achieve a positive impact. Compared with $\mathcal{F}$+$\mathcal{R}$ (w/ DN), $\mathcal{F}$+$\mathcal{S}$+$\mathcal{R}$ (w/ DN) improved the performance from 85.76\% to 87.93\% on Spearman's rank correlation via introducing procedure segmentation.
Compared with $\mathcal{F}$+$\mathcal{S}$+$\mathcal{R}$ (w/ DN), TSA (w/ DN) further improved the performance from 87.93\% to 92.03\% to on Spearman's rank correlation, which demonstrated the superiority of learning procedure-aware cross attention between query and exemplar procedures.

\begin{table}[t]
\caption{Effects of the number of exemplars for voting.}
\vspace{-5pt}
\footnotesize
\renewcommand\tabcolsep{10pt}
\label{table:vote_num}
\centering
\begin{tabular}{c|cc|c|c}
\toprule
\multirow{2}{*}{$M$} &\multicolumn{2}{c|}{AIoU@} & \multirow{2}{*}{$\rho$} & \multirow{2}{*}{R-$\ell_2(\times100)$}  \\
\cline{2-3}
& 0.5 & 0.75 & & \\
\midrule
1 &  76.01  &   26.56  & 0.9085  & 0.4020 \\
5  &  80.64  & 31.78 &  0.9154  & 0.3658 \\
\textbf{10} & \textbf{82.51} & \textbf{34.31} &  \textbf{0.9203} &  \textbf{0.3420}\\
15  &  82.52  & 34.31 &  0.9204  & 0.3419 \\
\bottomrule
\end{tabular}
\vspace{-10pt}
\end{table}

Besides, for the multi-exemplar voting used in inference, the number of exemplars $M$ is an important hyper-parameter that is a trade-off between better performance and larger computational costs. In Table \ref{table:vote_num}, we conducted some experiments to study the impact of $M$ on our approach TSA (w/ DN). It can be seen that with $M$ increasing, the performance becomes better while the computational cost is larger. The improvement on Spearman's rank correlation becomes less significant when $M\!>\!10$ and the similar trend on Relative $\ell_2$-distance also can be found in Table \ref{table:vote_num}.

\subsection{Visualization}
We visualize the procedure-aware cross-attention between query and exemplar on FineDiving, as shown in Figure \ref{visualization}. It can be seen that our approach highlights semantic, spatial, and temporal correspondent regions in the exemplar steps consistent with the query step, which makes the relative scores between query and exemplar procedures learned from fine-grained contrastive regression more interpretable.

\section{Conclusion and Discussion}
In this paper, we have constructed the first fine-grained sports video dataset, namely FineDiving, for assessing action quality. On FineDiving, we have proposed a procedure-aware action quality assessment approach via constructing a new temporal segmentation attention module, which learns semantic, spatial, and temporal consistent regions in pairwise steps in the query and exemplar procedures to make the inference process more interpretable, and achieve substantial improvements for existing AQA methods.

\textbf{Limitations \& Potential Negative Impact.} The proposed method has an assumption that the number of step transitions in the action procedure is known. The fine-grained annotations need to be manually decomposed and professionally labeled.

\textbf{Existing Assets and Personal Data.} This work contributes a new dataset on the diving sport, where all the data is collected and downloaded on YouTube and bilibili websites. We are actively contacting the creators to ensure that appropriate consent has been obtained.

%%%%%%%%%%%%%%%%%%%%%%%%%%%%%%%%%%%
\begin{appendix}

\section{More Experiment Settings}

\subsection{The Criterion of Selecting Exemplars}

For our approach TSA  (w/ DN), we selected exemplars from the training set based on the action types. In training, for each instance, we randomly selected an instance as an exemplar from the rest of the training instances with the same action type. In inference, we adopted the multi-exemplar voting strategy, i.e., randomly selecting 10 instances as 10 exemplars from the training instances with the same action type.
For our approach TSA  (w/o DN), we randomly selected exemplars from the training set in both training and inference.

\subsection{The Number of Step Transitions}

In FineDiving, the 60\% action types (e.g., 307C and 614B) consist of 3 sub-action types, corresponding to 3 steps; 30\% action types (e.g., 5251B and 6241B) consist of 4 sub-action types, corresponding to 4 steps; 10\% action types (e.g., 5152B and 5156B) consist of 4 sub-action types, corresponding to 5 steps. We provided full annotations of 5 possible steps to support future research on utilizing fine-grained annotations for AQA.
In experiments, our approach keeps $L$ constant and equals 2, indicating segmenting a dive action into 3 steps, i.e., 2 step transitions.
The reasons are as follows.
A dive action consists of \emph{take-off}, \emph{flight}, and \emph{entry}. If \emph{flight} is accomplished by one sub-action type, a dive action is divided into 3 steps, e.g., 307C containing 3 steps: Reverse, 3.5 Soms.Tuck, and Entry. If \emph{flight} is accomplished by two sub-action types successively, a dive action is divided into 4 steps, e.g., 6241B containing 4 steps: Arm.Back, 0.5 Twist, 2 Soms.Pike, and Entry. If \emph{flight} is accomplished by two sub-action types and one sub-action type is interspersed in another, a dive action is divided into 5 steps, e.g., 5152B containing 5 steps: Forward, 2.5 Soms.Pike, 1 Twist, 2.5 Soms.Pike, and Entry, where ``1 Twist'' is interspersed in ``2.5 Soms.Pike''.
We see that two cases of \emph{flight} accomplished by two sub-action types can be seen as two special cases of \emph{flight} accomplished by one sub-action type. Therefore, $L\!=\!2$ makes sense.

\section{More Ablation Study}

We summarize hyper-parameters for training as follows: the number of frames in each step is fixed into 5 before being fed into Multi-head Cross-Attention (MCA); all configurations are based on transformer decoder with 3 layers and 8 heads.

\begin{table}[ht]
\caption{Effects of the number of frames in each step in the procedure-aware cross-attention model.}
\vspace{-6pt}
\footnotesize
\renewcommand\tabcolsep{10pt}
\label{table:step_length}
\centering
\begin{tabular}{c|cc|c|c}
\toprule
\multirow{2}{*}{$L_{\text{step}}$} &\multicolumn{2}{c|}{AIoU@} & \multirow{2}{*}{$\rho$} & \multirow{2}{*}{R-$\ell_2(\times100)$}  \\
\cline{2-3}
& 0.5 & 0.75 & & \\
\midrule
3  & 76.49  & 24.17 &  0.9081  & 0.4003 \\
\textbf{5} & \textbf{82.51} & \textbf{34.31} &  \textbf{0.9203} &  \textbf{0.3420}\\
8  &  80.65  & 33.37 & 0.9198 & 0.3501 \\
10  &  77.97  & 30.44 & 0.9174 & 0.3588 \\
15  &  77.96  & 30.32 & 0.9149 & 0.3675 \\
\bottomrule
\end{tabular}
\end{table}

To investigate the effects of the number of frames in each step (denoted as $L_{\text{step}}$) on the performance of action quality assessment, we conduct several experiments on the FineDiving dataset. Table~\ref{table:step_length} summarizes the performance with different $L_{\text{step}}$ including 3, 5, 8, 10 and 15. We observe that when $L_{\text{step}}$ increases from 3 to 5, the action quality assessment performance achieves 1.22\% and 0.0583 improvements respectively on Spearman's rank correlation and Relative $\ell_2$-distance. When $L_{\text{step}}$ is bigger than 8, the performance tends to be flat with a slight decrease. Specifically, compared with the performance of $L_{\text{step}}$=5, that of $L_{\text{step}}$=8 degrades 0.05\% and 0.0081 on Spearman's rank correlation and Relative $\ell_2$-distance, respectively.
The experimental results illustrate that the number of frames in each step is not proportional to the action quality assessment performance, since each step containing fewer frames cannot make full use of intra-step information while each step containing too many frames may introduce some noisy information.

\begin{table}[ht]
\caption{Effects of the number of transformer decoder layers.}
\vspace{-6pt}
\footnotesize
\renewcommand\tabcolsep{10pt}
\label{table:layer_num}
\centering
\begin{tabular}{c|cc|c|c}
\toprule
\multirow{2}{*}{$R$} &\multicolumn{2}{c|}{AIoU@} & \multirow{2}{*}{$\rho$} & \multirow{2}{*}{R-$\ell_2(\times100)$}  \\
\cline{2-3}
& 0.5 & 0.75 & & \\
\midrule
1  & 81.71  & 33.24 &  0.9168  & 0.3612 \\
\textbf{3} & \textbf{82.51} & \textbf{34.31} &  \textbf{0.9203} &  \textbf{0.3420}\\
5  &  79.83  & 31.84 &  0.9202  & 0.3483 \\
10  &  79.57  & 31.78 &  0.9171  & 0.3534 \\
\bottomrule
\end{tabular}
\end{table}

To explore the effect of the number of transformer decoder layers (denoted as $R$) on the performance of action quality assessment, we conduct several experiments on the FineDiving dataset. Table~\ref{table:layer_num} summarizes the performance with different $R$, namely 1, 3, 5, and 10. It can be seen that when $R$ equals 3, the action quality assessment performance reaches the peaks (namely 0.9203 and 0.3420) on Spearman's rank correlation and Relative $\ell_2$-distance, respectively. The performance tends to be flat with a slight decrease when $R\!>$5, since too many transformer decoder layers may lead to overfitting for each step containing 5 frames ($L_{\text{step}}$=5).

\section{Code}

We provide the FineDiving dataset and code of our approach\footnote{\url{https://github.com/xujinglin/FineDiving}}, including the training and inference phases.

\begin{table*}
\caption{The detailed descriptions of action and sub-action types.}
\vspace{-5pt}
\footnotesize
\renewcommand\tabcolsep{6pt}
\renewcommand\arraystretch{0.9}
\label{dataset_info}
\centering
\begin{tabular}{c|c|c|c|c|c|c|c|c|c|c|c}
\toprule
\multirow{2}{*}{Action Type} & \multicolumn{5}{c|}{Sub-action Type} & \multirow{2}{*}{Action Type} & \multicolumn{5}{c}{Sub-action Type}\\
\cline{2-6}\cline{8-12}
& Take-off & \multicolumn{3}{c|}{Flight} & Entry & & Take-off & \multicolumn{3}{c|}{Flight} & Entry\\
\midrule
101B & Forward & \multicolumn{3}{c|}{0.5 Som.Pike} & \multirow{25}{*}{Entry} & 612B & Arm.Fwd & \multicolumn{3}{c|}{1 Som.Pike} & \multirow{5}{*}{Entry}\\
103B & Forward & \multicolumn{3}{c|}{1.5 Soms.Pike} & & 614B & Arm.Fwd & \multicolumn{3}{c|}{2 Soms.Pike} & \\
105B & Forward & \multicolumn{3}{c|}{2.5 Soms.Pike} & & 626B & Arm.Back & \multicolumn{3}{c|}{3 Soms.Pike} & \\
107B & Forward & \multicolumn{3}{c|}{3.5 Soms.Pike} & & 626C & Arm.Back & \multicolumn{3}{c|}{3 Soms.Tuck} & \\
109B & Forward & \multicolumn{3}{c|}{4.5 Soms.Pike} & & 636C & Arm.Reverse & \multicolumn{3}{c|}{3 Soms.Tuck} & \\\cline{7-12}
107C & Forward & \multicolumn{3}{c|}{3.5 Soms.Tuck} & & 5231D & Back & 0.5 Twist & \multicolumn{2}{c|}{1.5 Soms.Pike} & \multirow{16}{*}{Entry}\\
109C & Forward & \multicolumn{3}{c|}{4.5 Soms.Tuck} & & 5233D & Back & 1.5 Twists & \multicolumn{2}{c|}{1.5 Soms.Pike} & \\
201A & Back & \multicolumn{3}{c|}{0.5 Som.Straight} & & 5235D & Back & 2.5 Twists & \multicolumn{2}{c|}{1.5 Soms.Pike} & \\
201B & Back & \multicolumn{3}{c|}{0.5 Som.Pike} & & 5237D & Back & 3.5 Twists & \multicolumn{2}{c|}{1.5 Soms.Pike} & \\
201C & Back & \multicolumn{3}{c|}{0.5 Som.Tuck} & & 5251B & Back & 0.5 Twist & \multicolumn{2}{c|}{2.5 Soms.Pike} & \\
205B & Back & \multicolumn{3}{c|}{2.5 Soms.Pike} & & 5253B & Back & 1.5 Twists & \multicolumn{2}{c|}{2.5 Soms.Pike} & \\
207B & Back & \multicolumn{3}{c|}{3.5 Soms.Pike} & & 5255B & Back & 2.5 Twists & \multicolumn{2}{c|}{2.5 Soms.Pike} & \\
205C & Back & \multicolumn{3}{c|}{2.5 Soms.Tuck} & & 5331D & Reverse & 0.5 Twist & \multicolumn{2}{c|}{1.5 Soms.Pike} & \\
207C & Back & \multicolumn{3}{c|}{3.5 Soms.Tuck} & & 5335D & Reverse & 2.5 Twists & \multicolumn{2}{c|}{1.5 Soms.Pike} & \\
301B & Reverse & \multicolumn{3}{c|}{0.5 Som.Pike} & & 5337D & Reverse & 3.5 Twists & \multicolumn{2}{c|}{1.5 Soms.Pike} & \\
305B & Reverse & \multicolumn{3}{c|}{2.5 Soms.Pike} & & 5353B & Reverse & 1.5 Twists & \multicolumn{2}{c|}{2.5 Soms.Pike} & \\
303C & Reverse & \multicolumn{3}{c|}{1.5 Soms.Tuck} & & 5355B & Reverse & 2.5 Twists & \multicolumn{2}{c|}{2.5 Soms.Pike} & \\
305C & Reverse & \multicolumn{3}{c|}{2.5 Soms.Tuck} & & 6142D & Arm.Fwd & 1 Twist & \multicolumn{2}{c|}{2 Soms.Pike} & \\
307C & Reverse & \multicolumn{3}{c|}{3.5 Soms.Tuck} & & 6241B & Arm.Back & 0.5 Twist & \multicolumn{2}{c|}{2 Soms.Pike} & \\
401B & Inward & \multicolumn{3}{c|}{0.5 Som.Pike} & & 6243D & Arm.Back & 1.5 Twists & \multicolumn{2}{c|}{2 Soms.Pike} & \\
403B & Inward & \multicolumn{3}{c|}{1.5 Soms.Pike} & & 6245D & Arm.Back & 2.5 Twists & \multicolumn{2}{c|}{2 Soms.Pike} & \\ \cline{7-12}
405B & Inward & \multicolumn{3}{c|}{2.5 Soms.Pike} & & 5132D & Forward & 1.5 Soms.Pike & 1 Twist & 1.5 Soms.Pike & \multirow{5}{*}{Entry}\\
407B & Inward & \multicolumn{3}{c|}{3.5 Soms.Pike} & & 5152B & Forward & 2.5 Soms.Pike & 1 Twist & 2.5 Soms.Pike & \\
405C & Inward & \multicolumn{3}{c|}{2.5 Soms.Tuck} & & 5154B & Forward & 2.5 Soms.Pike & 2 Twists & 2.5 Soms.Pike &  \\
407C & Inward & \multicolumn{3}{c|}{3.5 Soms.Tuck} & & 5156B & Forward & 2.5 Soms.Pike & 3 Twists & 2.5 Soms.Pike &  \\
409C & Inward & \multicolumn{3}{c|}{4.5 Soms.Tuck} & & 5172B & Forward & 3.5 Soms.Pike & 1 Twist & 3.5 Soms.Pike &  \\
\bottomrule
\end{tabular}
\end{table*}

\section{The FineDiving Dataset Details}

We will release the FineDiving dataset to promote future research on action quality assessment.

\subsection{Descriptions of Action and Sub-action Types}

Table \ref{dataset_info} shows the detailed descriptions of action and sub-action types in FineDiving mentioned in the lexicon in subsection 3.1. Dataset Construction. We see that a combination of the sub-action types from three phases (namely take-off, flight, and entry) generates an action type. Specifically, the take-off phase is annotated by one of six sub-action types which are ``Forward, Back, Reverse, Inward, Armstand Forward, Armstand Back, and Armstand Reverse''. The entry phase is annotated by the sub-action type ``Entry''. The flight phase is labeled by one or two sub-action types describing the somersault process, where the number of sub-action types depends on the somersault process whether containing the twist or not.

As shown in Table \ref{dataset_info}, if the somersault process contains the twist, the flight phase is annotated by two sub-action types, where one sub-action type annotates the number of twist turns in the somersault process and another is annotated the number of somersault turns in the somersault process. The former sub-action type is a part of the latter sub-action type and does not be performed independently without the latter sub-action type. For different action types, the former sub-action type (twist) may occur at different locations in the somersault process. When the twist is performed at the beginning of the somersault process, we first annotate the number of twist turns and then annotate the number of somersault turns, such as the action types 5231D and 6245D in Table \ref{dataset_info}. When the twist is performed at the middle of the somersault process, we first annotate the number of somersault turns, then annotate the number of twist turns, and finally annotate the number of somersault turns, such as 5132D and 5172B in Table \ref{dataset_info}. Note that, the first and the last annotated the same somersault process but are separated by the twist.
If the somersault process does not contain the twist, the flight phase is annotated by one sub-action type, that is, the number of somersault turns, such as 205B and 626C in Table \ref{dataset_info}.

\subsection{Annotation Tool}

\begin{figure*}
  \centering
  \includegraphics[width=\linewidth]{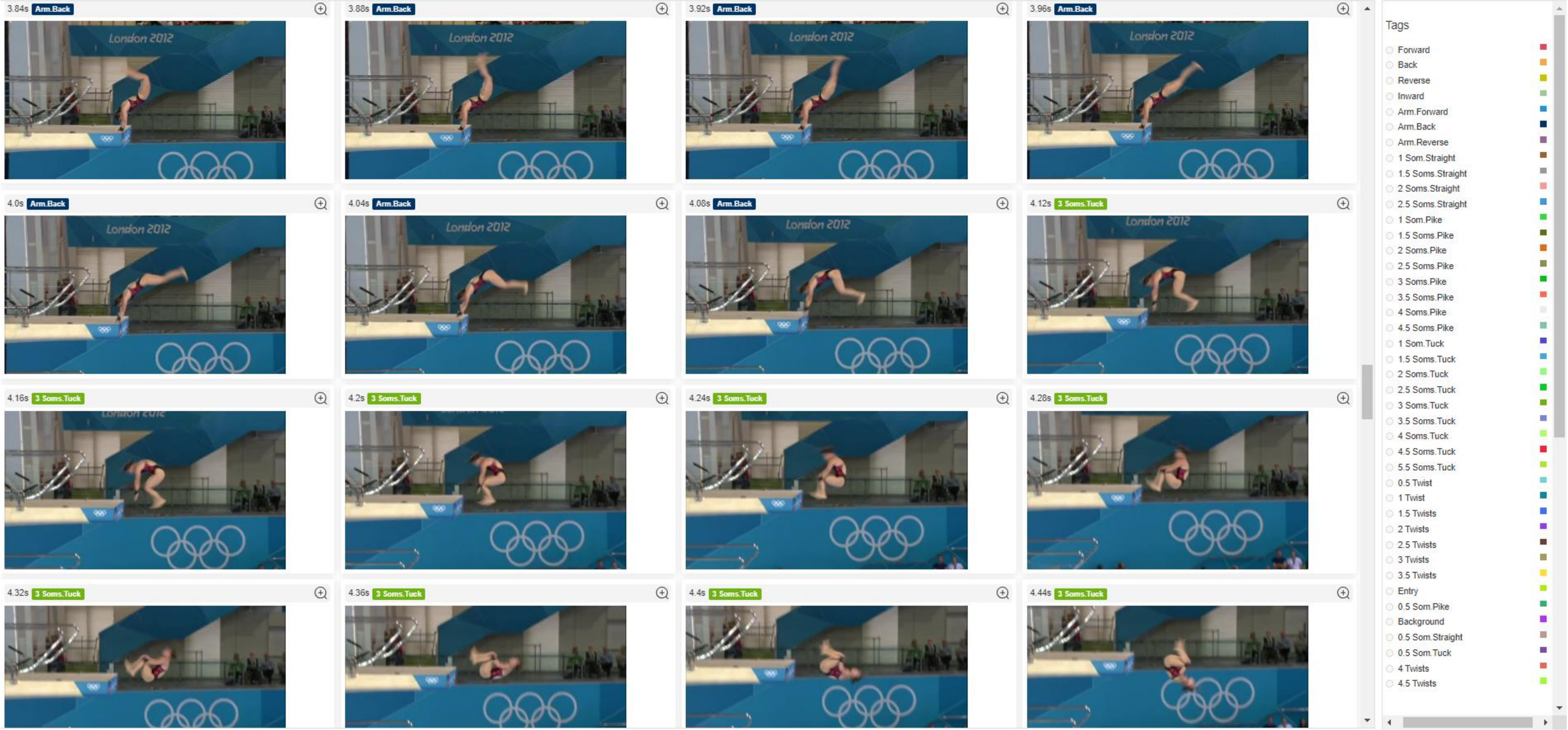}
  \caption{The interface for sub-action type annotation tool. The left part represents the video frames to be annotated by the pre-defined sub-action types. The right part shows the pre-defined sub-action types using different colors.}
  \vspace{-10pt}
  \label{tool}
\end{figure*}

In the fine-grained annotation stage, the durations of sub-action types are different for different action instances, which may cost a huge workload to label the FineDiving dataset with a conventional annotation tool. To improve the annotation efficiency, we utilize a publically available annotation toolbox \cite{tang2019coin} (mentioned in Annotation in subsection 3.1. Dataset Construction) to generate the frame-wise labels for various sub-action types, which can ensure high efficiency, accuracy, and consistency of our annotation results. Figure \ref{tool} shows an example interface of the annotation tool, which annotates the frames extracted from an action instance.

\end{appendix}

%%%%%%%%% REFERENCES
{\small
\bibliographystyle{ieee_fullname}
\bibliography{egbib}

\begin{thebibliography}{10}\itemsep=-1pt

\bibitem{bertasius2017baller}
Gedas Bertasius, Hyun Soo~Park, Stella~X Yu, and Jianbo Shi.
\newblock Am i a baller? basketball performance assessment from first-person
  videos.
\newblock In {\em ICCV}, pages 2177--2185, 2017.

\bibitem{carreira2017quo}
Joao Carreira and Andrew Zisserman.
\newblock Quo vadis, action recognition? a new model and the kinetics dataset.
\newblock In {\em CVPR}, pages 6299--6308, 2017.

\bibitem{chen2021sportscap}
Xin Chen, Anqi Pang, Wei Yang, Yuexin Ma, Lan Xu, and Jingyi Yu.
\newblock Sportscap: Monocular 3d human motion capture and fine-grained
  understanding in challenging sports videos.
\newblock {\em arXiv preprint arXiv:2104.11452}, 2021.

\bibitem{dosovitskiy2020image}
Alexey Dosovitskiy, Lucas Beyer, Alexander Kolesnikov, Dirk Weissenborn,
  Xiaohua Zhai, Thomas Unterthiner, Mostafa Dehghani, Matthias Minderer, Georg
  Heigold, Sylvain Gelly, et~al.
\newblock An image is worth 16x16 words: Transformers for image recognition at
  scale.
\newblock {\em arXiv preprint arXiv:2010.11929}, 2020.

\bibitem{doughty2018s}
Hazel Doughty, Dima Damen, and Walterio Mayol-Cuevas.
\newblock Who's better? who's best? pairwise deep ranking for skill
  determination.
\newblock In {\em CVPR}, pages 6057--6066, 2018.

\bibitem{doughty2019pros}
Hazel Doughty, Walterio Mayol-Cuevas, and Dima Damen.
\newblock The pros and cons: Rank-aware temporal attention for skill
  determination in long videos.
\newblock In {\em CVPR}, pages 7862--7871, 2019.

\bibitem{caba2015activitynet}
Bernard~Ghanem Fabian Caba~Heilbron, Victor~Escorcia and Juan~Carlos Niebles.
\newblock Activitynet: A large-scale video benchmark for human activity
  understanding.
\newblock In {\em CVPR}, pages 961--970, 2015.

\bibitem{feichtenhofer2016convolutional}
Christoph Feichtenhofer, Axel Pinz, and Andrew Zisserman.
\newblock Convolutional two-stream network fusion for video action recognition.
\newblock In {\em CVPR}, pages 1933--1941, 2016.

\bibitem{gattupalli2017cognilearn}
Srujana Gattupalli, Dylan Ebert, Michalis Papakostas, Fillia Makedon, and
  Vassilis Athitsos.
\newblock Cognilearn: A deep learning-based interface for cognitive behavior
  assessment.
\newblock In {\em IUI}, pages 577--587, 2017.

\bibitem{THUMOS15}
A. Gorban, H. Idrees, Y.-G. Jiang, A. Roshan~Zamir, I. Laptev, M. Shah, and R.
  Sukthankar.
\newblock {THUMOS} challenge: Action recognition with a large number of
  classes.
\newblock \url{http://www.thumos.info/}, 2015.

\bibitem{gu2018ava}
Chunhui Gu, Chen Sun, David~A Ross, Carl Vondrick, Caroline Pantofaru, Yeqing
  Li, Sudheendra Vijayanarasimhan, George Toderici, Susanna Ricco, Rahul
  Sukthankar, et~al.
\newblock Ava: A video dataset of spatio-temporally localized atomic visual
  actions.
\newblock In {\em CVPR}, pages 6047--6056, 2018.

\bibitem{hong2021video}
James Hong, Matthew Fisher, Micha{\"e}l Gharbi, and Kayvon Fatahalian.
\newblock Video pose distillation for few-shot, fine-grained sports action
  recognition.
\newblock In {\em ICCV}, pages 9254--9263, 2021.

\bibitem{ji20123d}
Shuiwang Ji, Wei Xu, Ming Yang, and Kai Yu.
\newblock 3d convolutional neural networks for human action recognition.
\newblock {\em TPAMI}, 35(1):221--231, 2012.

\bibitem{karpathy2014large}
Andrej Karpathy, George Toderici, Sanketh Shetty, Thomas Leung, Rahul
  Sukthankar, and Li Fei-Fei.
\newblock Large-scale video classification with convolutional neural networks.
\newblock In {\em CVPR}, pages 1725--1732, 2014.

\bibitem{kingma2014adam}
Diederik~P Kingma and Jimmy Ba.
\newblock Adam: A method for stochastic optimization.
\newblock {\em arXiv preprint arXiv:1412.6980}, 2014.

\bibitem{kuehne2011hmdb}
Hildegard Kuehne, Hueihan Jhuang, Est{\'\i}baliz Garrote, Tomaso Poggio, and
  Thomas Serre.
\newblock Hmdb: a large video database for human motion recognition.
\newblock In {\em ICCV}, pages 2556--2563, 2011.

\bibitem{li2019action}
Hongyang Li, Jun Chen, Ruimin Hu, Mei Yu, Huafeng Chen, and Zengmin Xu.
\newblock Action recognition using visual attention with reinforcement
  learning.
\newblock In {\em ICMM}, pages 365--376, 2019.

\bibitem{li2018end}
Yongjun Li, Xiujuan Chai, and Xilin Chen.
\newblock End-to-end learning for action quality assessment.
\newblock In {\em PRCM}, pages 125--134, 2018.

\bibitem{li2021multisports}
Yixuan Li, Lei Chen, Runyu He, Zhenzhi Wang, Gangshan Wu, and Limin Wang.
\newblock Multisports: A multi-person video dataset of spatio-temporally
  localized sports actions.
\newblock In {\em ICCV}, pages 13536--13545, 2021.

\bibitem{li2018resound}
Yingwei Li, Yi Li, and Nuno Vasconcelos.
\newblock Resound: Towards action recognition without representation bias.
\newblock In {\em ECCV}, pages 513--528, 2018.

\bibitem{lin2019bmn}
Tianwei Lin, Xiao Liu, Xin Li, Errui Ding, and Shilei Wen.
\newblock Bmn: Boundary-matching network for temporal action proposal
  generation.
\newblock In {\em ICCV}, pages 3889--3898, 2019.

\bibitem{meyer2011assessing}
Meredith Meyer, Dare~A Baldwin, and Kara Sage.
\newblock Assessing young children's hierarchical action segmentation.
\newblock In {\em CogSci}, volume~33, 2011.

\bibitem{monfortmoments}
Mathew Monfort, Alex Andonian, Bolei Zhou, Kandan Ramakrishnan, Sarah~Adel
  Bargal, Tom Yan, Lisa Brown, Quanfu Fan, Dan Gutfruend, Carl Vondrick, et~al.
\newblock Moments in time dataset: one million videos for event understanding.
\newblock {\em TPAMI}, pages 1--8, 2019.

\bibitem{montes2016temporal}
Alberto Montes, Amaia Salvador, Santiago Pascual, and Xavier Giro-i Nieto.
\newblock Temporal activity detection in untrimmed videos with recurrent neural
  networks.
\newblock {\em arXiv preprint arXiv:1608.08128}, 2016.

\bibitem{niebles2010modeling}
Juan~Carlos Niebles, Chih-Wei Chen, and Li Fei-Fei.
\newblock Modeling temporal structure of decomposable motion segments for
  activity classification.
\newblock In {\em ECCV}, pages 392--405, 2010.

\bibitem{oneata2013action}
Dan Oneata, Jakob Verbeek, and Cordelia Schmid.
\newblock Action and event recognition with fisher vectors on a compact feature
  set.
\newblock In {\em ICCV}, pages 1817--1824, 2013.

\bibitem{pan2019action}
Jia-Hui Pan, Jibin Gao, and Wei-Shi Zheng.
\newblock Action assessment by joint relation graphs.
\newblock In {\em ICCV}, pages 6331--6340, 2019.

\bibitem{parmar2019action}
Paritosh Parmar and Brendan Morris.
\newblock Action quality assessment across multiple actions.
\newblock In {\em WACV}, pages 1468--1476, 2019.

\bibitem{parmar2017learning}
Paritosh Parmar and Brendan Tran~Morris.
\newblock Learning to score olympic events.
\newblock In {\em CVPRW}, pages 20--28, 2017.

\bibitem{mtlaqa}
Paritosh Parmar and Brendan Tran~Morris.
\newblock What and how well you performed? a multitask learning approach to
  action quality assessment.
\newblock In {\em CVPR}, pages 304--313, 2019.

\bibitem{pirsiavash2014assessing}
Hamed Pirsiavash, Carl Vondrick, and Antonio Torralba.
\newblock Assessing the quality of actions.
\newblock In {\em ECCV}, pages 556--571, 2014.

\bibitem{schmidt1976understanding}
Charles~F Schmidt.
\newblock Understanding human action: Recognizing the plans and motives of
  other persons.
\newblock 1976.

\bibitem{schuldt2004recognizing}
Christian Schuldt, Ivan Laptev, and Barbara Caputo.
\newblock Recognizing human actions: a local svm approach.
\newblock In {\em ICPR}, pages 32--36, 2004.

\bibitem{shao2020finegym}
Dian Shao, Yue Zhao, Bo Dai, and Dahua Lin.
\newblock Finegym: A hierarchical video dataset for fine-grained action
  understanding.
\newblock In {\em CVPR}, pages 2616--2625, 2020.

\bibitem{shao2020intra}
Dian Shao, Yue Zhao, Bo Dai, and Dahua Lin.
\newblock Intra-and inter-action understanding via temporal action parsing.
\newblock In {\em CVPR}, pages 730--739, 2020.

\bibitem{simonyan2014two}
Karen Simonyan and Andrew Zisserman.
\newblock Two-stream convolutional networks for action recognition in videos.
\newblock {\em arXiv preprint arXiv:1406.2199}, 2014.

\bibitem{soomro2012ucf101}
Khurram Soomro, Amir~Roshan Zamir, and Mubarak Shah.
\newblock Ucf101: A dataset of 101 human actions classes from videos in the
  wild.
\newblock {\em arXiv preprint arXiv:1212.0402}, 2012.

\bibitem{tang2019coin}
Yansong Tang, Dajun Ding, Yongming Rao, Yu Zheng, Danyang Zhang, Lili Zhao,
  Jiwen Lu, and Jie Zhou.
\newblock Coin: A large-scale dataset for comprehensive instructional video
  analysis.
\newblock In {\em CVPR}, pages 1207--1216, 2019.

\bibitem{tang2020uncertainty}
Yansong Tang, Zanlin Ni, Jiahuan Zhou, Danyang Zhang, Jiwen Lu, Ying Wu, and
  Jie Zhou.
\newblock Uncertainty-aware score distribution learning for action quality
  assessment.
\newblock In {\em CVPR}, pages 9839--9848, 2020.

\bibitem{tran2015learning}
Du Tran, Lubomir Bourdev, Rob Fergus, Lorenzo Torresani, and Manohar Paluri.
\newblock Learning spatiotemporal features with 3d convolutional networks.
\newblock In {\em ICCV}, pages 4489--4497, 2015.

\bibitem{varol2017long}
G{\"u}l Varol, Ivan Laptev, and Cordelia Schmid.
\newblock Long-term temporal convolutions for action recognition.
\newblock {\em TPAMI}, 40(6):1510--1517, 2017.

\bibitem{wang2013action}
Heng Wang and Cordelia Schmid.
\newblock Action recognition with improved trajectories.
\newblock In {\em ICCV}, pages 3551--3558, 2013.

\bibitem{wang2018non}
Xiaolong Wang, Ross Girshick, Abhinav Gupta, and Kaiming He.
\newblock Non-local neural networks.
\newblock In {\em CVPR}, pages 7794--7803, 2018.

\bibitem{xu2019learning}
Chengming Xu, Yanwei Fu, Bing Zhang, Zitian Chen, Yu-Gang Jiang, and Xiangyang
  Xue.
\newblock Learning to score figure skating sport videos.
\newblock {\em TCSVT}, 30(12):4578--4590, 2019.

\bibitem{yang2020temporal}
Ceyuan Yang, Yinghao Xu, Jianping Shi, Bo Dai, and Bolei Zhou.
\newblock Temporal pyramid network for action recognition.
\newblock In {\em CVPR}, pages 591--600, 2020.

\bibitem{yeung2016end}
Serena Yeung, Olga Russakovsky, Greg Mori, and Li Fei-Fei.
\newblock End-to-end learning of action detection from frame glimpses in
  videos.
\newblock In {\em CVPR}, pages 2678--2687, 2016.

\bibitem{yu2021group}
Xumin Yu, Yongming Rao, Wenliang Zhao, Jiwen Lu, and Jie Zhou.
\newblock Group-aware contrastive regression for action quality assessment.
\newblock In {\em ICCV}, pages 7919--7928, 2021.

\bibitem{zhang2014relative}
Qiang Zhang and Baoxin Li.
\newblock Relative hidden markov models for video-based evaluation of motion
  skills in surgical training.
\newblock {\em TPAMI}, 37(6):1206--1218, 2014.

\bibitem{zhao2019hacs}
Hang Zhao, Antonio Torralba, Lorenzo Torresani, and Zhicheng Yan.
\newblock Hacs: Human action clips and segments dataset for recognition and
  temporal localization.
\newblock In {\em ICCV}, pages 8668--8678, 2019.

\bibitem{zhao2017temporal}
Yue Zhao, Yuanjun Xiong, Limin Wang, Zhirong Wu, Xiaoou Tang, and Dahua Lin.
\newblock Temporal action detection with structured segment networks.
\newblock In {\em ICCV}, pages 2914--2923, 2017.

\end{thebibliography}
}

\end{document}